\documentclass[lettersize,journal]{IEEEtran}

\usepackage{amsmath,amsfonts}
\usepackage{algorithmic}
\usepackage{algorithm}
\usepackage{array}
\usepackage{textcomp}
\usepackage{stfloats}
\usepackage{url}
\usepackage{verbatim}
\usepackage{graphicx}
\usepackage{cite}

\usepackage{color}

\usepackage{booktabs}
\usepackage{dsfont}
\usepackage{color}
\usepackage{multirow}
\usepackage{forest}
\usepackage{amsmath}
\usepackage{amsfonts}
\usepackage{graphics}
\usepackage{graphicx}

\usepackage{subcaption}
\usepackage{layouts}

\usepackage{caption}
\usepackage{mwe}
\usepackage{amssymb}
\usepackage{bm}
\usepackage{makecell}
\usepackage{tabularx}
\usepackage{arydshln}
\usepackage{url}
\usepackage{CJKutf8}
\usepackage{wrapfig}
\usepackage{enumitem}
\usepackage{diagbox}
\usepackage{ascii}

\definecolor{nred}{RGB}{196, 38, 11}
\definecolor{nblue}{RGB}{41, 52, 190}
\definecolor{ngreen}{RGB}{18, 141, 21}

\definecolor{zptu}{RGB}{18, 141, 21}

\begin{document}

\title{Understanding and Mitigating the Uncertainty in Zero-Shot Translation}

\author{Wenxuan~Wang,
        Wenxiang~Jiao,
        Shuo~Wang,
        Zhaopeng~Tu,
        and Michael~R.~Lyu,~\IEEEmembership{Fellow,~IEEE}%
}

\markboth{Journal of \LaTeX\ Class Files,~Vol.~14, No.~8, August~2021}%
{Shell \MakeLowercase{\textit{et al.}}: A Sample Article Using IEEEtran.cls for IEEE Journals}


\maketitle

\begin{abstract}
Zero-shot translation is a promising direction for building a comprehensive multilingual neural machine translation~(MNMT) system. However, its quality is still not satisfactory due to off-target issues. In this paper, we aim to understand and alleviate the off-target issues from the perspective of uncertainty in zero-shot translation. By carefully examining the translation output and model confidence, we identify two uncertainties that are responsible for the off-target issues, namely, extrinsic data uncertainty and intrinsic model uncertainty.
Based on the observations, we propose two lightweight and complementary approaches to denoise the training data for model training and explicitly penalize the off-target translations by unlikelihood training during model training. 
Extensive experiments on both balanced and imbalanced datasets show that our approaches significantly improve the performance of zero-shot translation over strong MNMT baselines. 
\end{abstract}


\section{Introduction}

\IEEEPARstart{M}{ultilingual} neural machine translation~(MNMT) aims to translate between any two languages with a unified model~\cite{Johnson:2017:TACL,Aharoni:2019:NAACL,Wang2022SynchronousIF,jiao2022tencent}.
It is appealing due to the model's efficiency, easy deployment, and knowledge transfer between languages. 
Previous studies~\cite{Johnson:2017:TACL,Gu:2019:ACL,hou2022adapters} suggest that knowledge transfer in MNMT significantly improves the performance of low-resource translation, and has the potential for zero-shot translation between language pairs unseen in the training process. 
For example, a widely-adopted setting is that the training data contains non-English to English~(X-En) and English to non-English (En-X) sentence pairs~\cite{Johnson:2017:TACL, Zhang2020ACL, Tang2020MultilingualTW, Yang2021ImprovingMT, Wang:2021:EMNLP}. 
The MNMT model trained on such data can conduct zero-shot translation between two non-English languages (X-X).
Since it is costly and even unrealistic to build parallel data for all language pairs, improving the quality of zero-shot translation is a promising direction for developing a comprehensive and well-performing MNMT system.

However, zero-shot translation suffers from serious {\bf off-target issues}~\cite{Gu:2019:ACL,Zhang2020ACL,Chen2023OnTO}, where the MNMT model tends to translate into other languages rather than the expected target language. Table~\ref{tab:examples-off-target} shows off-target examples for zero-shot translation.
As a result, the quality of zero-shot translation is far from satisfactory for practical application. 

A number of recent efforts have explored ways to improve zero-shot translation by mitigating off-target issues. One thread of work focuses on modifying the model architecture~\cite{Zhang2020ACL,Liu2021ImprovingZT,Wu2021LanguageTM} or introducing auxiliary training losses~\cite{Arivazhagan2019MassivelyMN,AlShedivat2019ConsistencyBA,Yang2021ImprovingMT} to enhance the flexible translation relations in MNMT. Another thread of work aims to generate synthetic data for zero-shot translation pairs in either an offline~\cite{Gu:2019:ACL} or online~\cite{Zhang2020ACL} manner. 
These approaches require additional efforts for model modification and computational costs.

\begin{table}[t!]
\centering
\caption{Off-target examples for zero-shot Fr-De in multilingual translation. The MNMT model often {\color{nred} copies} the source sentence or translates into {\color{nblue}other languages} (e.g. English rather than German).}
\begin{tabular}{l m{5cm} }
\toprule
\bf Category &\bf \qquad\qquad Examples  \\
\midrule
\multirow{3}{*}{to-Source} &
 {\it src} (\texttt{Fr}): Tout semble s'être bien passé. \\
 &{\it tgt} (\texttt{De}): Alles scheint gut gelaufen zu sein. \\
 &{\it hyp} (\texttt{Fr}): \color{nred} Tout semble s'être bien passé.  \\
\midrule
\multirow{3}{*}{to-Others} &  {\it src} (\texttt{Fr}): Mettez-le sur quatre assiettes.  \\
  &  {\it tgt} (\texttt{De}): Legen Sie ihn auf vier Teller.  \\
  &  {\it hyp} (\texttt{En}): \color{nblue} Put it on four plates.  \\
\bottomrule
\end{tabular}
\label{tab:examples-off-target}
\end{table}

In this work, we aim to better understand and mitigate off-target issues in zero-shot translation.
We first empirically connect the widely-cited off-target issues in zero-shot translation to the uncertain prediction of MNMT models, which assign high confidence to the off-target translations for zero-shot language pairs (\S~\ref{sec_ana}).
In uncertainty theory, there are different sources of uncertainty for engineering systems. These uncertainties can come from model design,  named intrinsic uncertainties, or from model calibrations using uncertain data, named extrinsic uncertainty~\cite{Steck1989ExperimentationAU}.
Following this framework, we then identify two language uncertainties that are responsible for the uncertain prediction of target languages. Extrinsic uncertainty is the uncertainty caused outside the model while intrinsic uncertainty is the uncertainty caused by model design:
\begin{itemize}[leftmargin=10pt]
    \item {\bf extrinsic data uncertainty} (\S~\ref{sec:data-uncertainty}): we show that for 5.8\% of the training examples in the commonly used multilingual data OPUS~\cite{Zhang2020ACL}, the target sentences are in the source language. Previous studies have shown that such data noises~\cite{Ott2018ICML,jiao2020data,jiao2022exploiting} can significantly affect the model uncertainty for bilingual NMT. Our study empirically reconfirms these findings for zero-shot translation, which is more sensitive to the data noises without supervision from parallel data.
    \item {\bf intrinsic model uncertainty} (\S~\ref{sec:model-uncertainty}): we show that MNMT models tend to spread too much probability mass over the vocabulary of off-target languages in zero-shot translation, resulting in an overall over-estimation of hypotheses in off-target languages. In contrast, the trend does not hold for supervised translations. 
\end{itemize}

Starting from the above observations, we propose two lightweight and complementary approaches to mitigate the data and model uncertainties. 
For data uncertainty, we remove the off-target sentence pairs from training data to make sure the MNMT models can learn more correct mappings between languages during training. For model uncertainty, we propose unlikelihood training to explicitly penalize the off-target translations in training, which can perform better when the counteractive effect of data uncertainty is removed.

Experimental results across different MNMT scenarios show that our approaches significantly improve zero-shot translation performance over strong MNMT baselines.
Extensive analyses demonstrate that our approaches successfully reduce the ratios of off-target translations from more than 20\% to as low as 1.1\%.

\vspace{3pt}\noindent{\bf \em Contributions} The main contributions of our work are listed as follows:
\begin{itemize}[leftmargin=10pt]
    \item We identify two uncertainties, namely extrinsic data uncertainty and intrinsic model uncertainty, which are responsible for the off-target issues in zero-shot translation.
    \item We propose two effective approaches to mitigate the off-target issues, which introduce no or only marginal additional computational cost.
\end{itemize}

The rest of the paper is organized as follows:
We first introduce the background of multilingual translation and off-target translation, existing working, as well as the experimental setup in Section~\ref{sec-backgound}; Then in Section~\ref{sec_ana}, we present poor zero-shot performance of well-trained MNMT models due to off-target issues; In Section~\ref{sec:data-uncertainty} and Section~\ref{sec:model-uncertainty}, we identify two uncertainties that are responsible for the off-target issues, i.e., extrinsic data uncertainty and intrinsic model uncertainty, as well as the corresponding mitigating methods; Next, in Section~\ref{sec:main}, we empirically evaluate the effectiveness of the proposed methods; Finally, we review the previous work related to ours in Section~\ref{sec-related}.

\section{Preliminary}
\label{sec-backgound}

\subsection{Multilingual Neural Machine Translation}

Multilingual neural machine translation~(MNMT) aims to translate between any two languages with a unified translation model.
Early studies~\cite{Dong2015ACL} on MNMT follow a multi-task learning scheme, such as learning one-to-many translations with a shared encoder and separate decoders or many-to-one translations with separate encoders and a shared encoder. This kind of method impedes the knowledge transfer between languages and is also very inefficient in model storage when the languages scale. 
Later, \cite{Johnson:2017:TACL} successfully realized MNMT with a single model, achieving competitive performance with individual bilingual NMT models. 
It also enables zero-shot translation, which translates between language pairs unseen in the training process. 
For example, the training data contains non-English to English~(X-En) and English to non-English (En-X) sentence pairs. 
The MNMT model trained on such data can conduct zero-shot translation between two non-English languages (X-X).
In this paper, we follow this unified architecture to study the zero-shot translation of MNMT models.

\subsection{Definition of Off-Target Issue}

\textbf{Off-Target Issue} is a type of translation error that commonly occurs in zero-shot translation~\cite{Zhang2020ACL}. It describes the phenomenon that MNMT models ignore the given target language information and translate the source sentence into the wrong language.
Assume that $\mathcal{L}$ denotes the set of languages involved in the MNMT model, and $T\in\mathcal{L}$ is the target language, the off-target ratio (OTR) is calculated as:
\begin{align}
    \mathrm{\bf OTR} = \frac{\sum_{i=1}^N \mathds{1}_{l(\Tilde{y}_i) \neq T}}{N},
\end{align}
where $N$ is the number of test samples, and $l(\Tilde{y}_i)$ denotes the detected language of the translation $\Tilde{y}_i$.
We adopt OTR as one of the metrics to evaluate the performance of zero-shot translation in this work.

\subsection{Existing Works on Off-Target Issue}

A number of recent efforts have explored ways to improve zero-shot translation by mitigating the off-target issue. One thread of work focuses on modifying the model architecture, such as adding a target language-aware linear transformation between the encoder and the decoder~\cite{Zhang2020ACL} and removing residual connections in an encoder layer~\cite{Liu2021ImprovingZT}. Another thread of work introduces auxiliary tasks with additional training losses to help the model training, such as a likelihood training objective that encourages the model to produce equivalent translations of parallel sentences in auxiliary languages~\cite{AlShedivat2019ConsistencyBA}, an auxiliary target language prediction task to regularize decoder outputs to retain information about the target language~\cite{Yang2021ImprovingMT}, and an additional denoising autoencoder objective~\cite{Wang:2021:EMNLP}. However, the effectiveness of these methods is limited. The off-target ratios are still high according to our experiments.

Besides, researchers also try to generate synthetic data for zero-shot translation pairs in either an offline~\cite{Gu:2019:ACL} or online~\cite{Zhang2020ACL} manner. However, such methods require additional computational costs in generating data and model training. Also, adding data pairs in the zero-shot translation direction could hurt the performance of supervised translation, which is known as the curse of representation bottleneck in the multilingual translation field\cite{Zhang2020ACL}. 

In this paper, we will identify two uncertainties that are responsible for the off-target issues, extrinsic data uncertainty, i.e., for a portion of the training examples, the target sentences are in the source languages, and intrinsic model uncertainty, i.e., MNMT models tend to spread too much probability mass over the vocabulary of off-target languages.  Based on these, we provide simple and effective solutions, data denoising and unlikelihood training, to mitigate the data and model uncertainty, respectively.

\subsection{Experimental Setup}
In this section, we introduce the experimental setup throughout this work, including the training data, evaluation data, and model configurations.

\vspace{3pt}
\subsubsection{\bf\em Training Data}
We mainly conduct experiments on three  datasets across different data distributions and corpus sizes:
\begin{itemize}
    \item {\bf \em OPUS-100 Data} is an {\bf unbalanced} multilingual dataset, where some language pairs have more training instances than others. \cite{Zhang2020ACL} propose OPUS-100 which consists of 55M English-centric sentence pairs covering 100 languages. We also choose five language pairs from OPUS-100, including English-German (En-De),
    English-Chinese (En-Zh), English-Japanese (En-Ja), English-French (En-Fr), and English-Romanian (En-Ro) to construct {\bf \em balanced OPUS-6 Data} (5M in total with 1M each). 
    We follow~\cite{Zhang2020ACL} to apply BPE~\cite{Sennrich2016ACL} to learn a joint vocabulary size of 64K from the whole OPUS-100 dataset.
    \item {\bf \em WMT-6 Data} is a large-scale {\bf unbalanced} dataset. Specifically, we collect the language pairs same as OPUS-6 from the widely-used WMT competition dataset~\cite{Sennrich2016EdinburghNM,Chen2022IntegratingPT, Chen2020TowardsMD, Liu2019LatentAB, Yang2022GTransGA}, including WMT20 En-De (45.2M), WMT20 En-Zh (19.0M), WMT20 En-Ja (11.5M), WMT14 En-Fr (35.5M), and WMT16 En-Ro (0.6M).
    We learn a joint BPE~\cite{Sennrich2016ACL} model with 32K merge operations.
\end{itemize}

\subsubsection{\bf\em Validation Set}
For OPUS data, we use the original validation data provided by OPUS-100 as the validation set. For WMT data, we use WMT 19 En-De test set, WMT 19 En-Zh test set, WMT 20 En-Ja dev set, WMT14 En-Fr dev set, and WMT16 En-Ro dev set as validation set.

\vspace{3pt}
\subsubsection{\bf\em Multi-Source Test Set}
To eliminate the content bias across languages~\cite{Wang:2021:ACL}, we evaluate the performance of multilingual translation models on the multi-source TED58 test set~\cite{qi2018and,tran2020cross}, where each sentence is translated into multiple languages. We select the above six languages and filter the original test set to ensure that each sentence has translations in all six languages. Finally, we obtain 3804 sentences in six languages, i.e., 22824 sentences in total. We use the filtered test set to evaluate the performance on both supervised and zero-shot translations.
 We report the results of both BLEU scores~\cite{Papineni2002ACL} and off-target ratios (OTR) for both supervised and zero-shot translation. 
For example, the supervised translation and the zero-shot translation performance on OPUS-6 dataset are the average of 10 supervised directions (i.e., En-X and X-En) and 20 zero-shot directions (i.e., $X_i$-$X_j$), respectively. We employ the \texttt{langid} library\footnote{\url{https://github.com/saffsd/langid.py}}, the most widely used language identification tool with 93.7\% accuracy on 7 datasets across 97 languages, to detect the language of sentences and calculate the off-target ratio for zero-shot translation directions.
We also adopt two widely used evaluation metrics, COMET~\cite{Rei2020COMETAN} and chrF~\cite{Popovic2015chrFCN} to validate our method.

\vspace{3pt}
\subsubsection{\bf\em Model}
{
We adopt the Fairseq\footnote{\url{https://github.com/pytorch/fairseq}} toolkit for experiments. }
All NMT models in this paper follow the Transformer-big settings, with 6 layers, 1024 hidden sizes, and 16 heads. 
{
To distinguish languages, we add language tokens to the training samples by two strategies implemented in Fairseq, i.e., \textsc{S-Enc-T-Dec} and \textsc{T-Enc}.
The \textsc{S-Enc-T-Dec} strategy adds source language tokens at the encoder and target language tokens at the decoder, while T-Enc only adds target language tokens at the encoder.
}
For multilingual translation models, we train a Transformer-big model with 1840K tokens per batch for 50K updates. 
We conduct the experiments on 16 NVIDIA V100 GPUs and select the model by the lowest loss on the validation set.

\section{Analyzing Uncertainty}
\label{sec_ana}

In this section, we present the poor zero-shot performance of our well-trained MNMT models due to off-target issues. Then we link the off-target issues to the uncertain prediction of target languages.

\begin{table}[t!]
\centering
\caption{BLEU scores of bilingual and multilingual \textsc{Transformer-Big} models trained on {\bf OPUS-6} data for supervised translation. We report results on both the test sets provided by the OPUS data (``OPUS'') and the multi-source TED test set used in this work (``TED'').}
\begin{tabular}{c cc cc}
\toprule
\multirow{2}{*}{\bf Model}  &  \multicolumn{2}{c}{\bf English$\Rightarrow$X}  &  \multicolumn{2}{c}{\bf X$\Rightarrow$English}\\
 \cmidrule(lr){2-3} \cmidrule(lr){4-5}
  &  \bf OPUS  &  \bf TED  &  \bf OPUS  &  \bf TED\\
\midrule
Bilingual Model             &  32.0  &21.6   &  31.0 & 22.9\\
\midrule
\multicolumn{5}{c}{\bf Multilingual NMT Models}\\
\textsc{S-Enc-T-Dec}  &  34.8 & 26.9  & 33.5  & 27.2\\
\textsc{T-Enc}        &  34.8 & 27.1  & 33.5 & 27.4\\
\bottomrule
\end{tabular}
\label{tab:opus-preliminary}
\end{table}

\subsection{\bf\em MNMT Models are Well Trained}
In Table~\ref{tab:opus-preliminary}, we list the supervised translation performance of our multilingual NMT models, which are evaluated on both the OPUS test sets and the multi-source TED test set. 
For comparison, we also include the bilingual model for each language pair as baselines. For bilingual models on high-resource datasets, we train a Transformer-big model with 460K tokens per batch for 30K updates. As for low-resource datasets, we train a Transformer-base model with 16K tokens per batch for 50K updates.
Clearly, our MNMT models consistently and significantly outperform their bilingual counterparts, demonstrating that our models are well-trained so that the findings and improvement in this work are convincing.

\begin{table}[t]
\centering
\caption{BLEU scores and off-target ratios (OTR) of MNMT models on supervised and zero-shot test sets.
}
\begin{tabular}{c rr rr}
\toprule
\bf Training &  \multicolumn{2}{c}{\bf Supervised}  & \multicolumn{2}{c}{\bf Zero-Shot}\\
\cmidrule(lr){2-3}\cmidrule(lr){4-5}
\bf Data    &   \bf BLEU$\uparrow$  & \bf OTR$\downarrow$    &   \bf BLEU$\uparrow$  & \bf OTR$\downarrow$\\
\midrule
\multicolumn{5}{c}{\bf S-\textsc{Enc}-T-\textsc{Dec} Models}\\
OPUS-6   &  27.1 & 1.9 & 12.3 & 20.6 \\
WMT-6    & 28.0  & 1.8  & 10.6 &  37.8\\
\midrule
\multicolumn{5}{c}{\bf T-\textsc{Enc} Models}\\
OPUS-6   &  27.2 & 1.9 & 10.2 & 32.1\\
WMT-6    &  28.8 & 1.7 & 13.3 & 22.5\\
\bottomrule
\end{tabular}
\label{tab:vanilla-all}
\end{table}

\begin{table}[t]
\centering
\caption{BLEU and Off-target Ratios (OTR) on supervised translation(``Sup.'') and zero-shot translation(``Zero'') for \textsc{T-Enc} model trained on OPUS-6 and {WMT-6} data.}
\begin{tabular}{c rrr rrr}
\toprule
\multirow{2}{*}{\bf Target} &  \multicolumn{3}{c}{\bf \bf BLEU$\uparrow$}  & \multicolumn{3}{c}{\bf OTR$\downarrow$}\\
\cmidrule(lr){2-4}\cmidrule(lr){5-7}
    &   \bf Sup.   &  \bf Zero   & \bf $\triangle$    &   \bf Sup.   &  \bf Zero   & \bf $\triangle$\\
\midrule
\multicolumn{7}{c}{{\bf OPUS-6}}\\
Ja  & 19.0 & 15.7 & -3.3 & 0.4 & 2.1 & +1.7\\
Zh  & 23.1 & 11.4 & -11.7 & 0.4 & 32.6 & +32.2\\
De  & 29.4 & 6.1 & -23.3 & 2.8  & 49.6 & +46.8\\
Fr  & 37.1 & 6.4 & -30.7 & 2.7  & 61.8 & +59.1\\
Ro  & 26.8 & 11.4 & -15.4 & 3.7 & 14.3 & +10.6\\
\midrule
\multicolumn{7}{c}{{\bf WMT-6}}\\
Ja  & 26.1 & 18.1 & -8.0 & 0.8 & 2.0 & +1.2\\
Zh  & 26.4 & 17.1 & -9.3 & 0.9 & 6.7 & +5.8\\
De  & 33.2 & 8.1 & -25.1 & 1.9  & 41.8 & +39.9\\
Fr  & 38.9 & 12.3 & -26.6 & 1.7  & 39.0 & +37.3\\
Ro  & 24.1 & 11.0 & -13.1 & 4.8 & 23.0 & +18.2\\
\bottomrule
\end{tabular}
\label{tab:vanilla-case}
\end{table}

\subsection{\bf\em Poor Zero-Shot Performance and Off-Target Issues}

Table~\ref{tab:vanilla-all} lists the translation results. Compared with the supervised translation, the MNMT models produce lower-quality zero-shot translations (e.g., 15+ BLEU scores lower) due to much higher ratios of off-target translations (e.g., 32.1 vs. 1.9 on OPUS-6 {with \textsc{T-Enc}}).
To further validate our claim, we list the detailed results in Table~\ref{tab:vanilla-case}. As seen, the gap in BLEU score between supervised and zero-shot translations is highly correlated to that of OTR, showing the high correlation between translation performance and off-target issues.



\begin{figure}[t]
    \centering 
    \subfloat[Supervised Fr-En]{
    \includegraphics[height=0.3\textwidth]{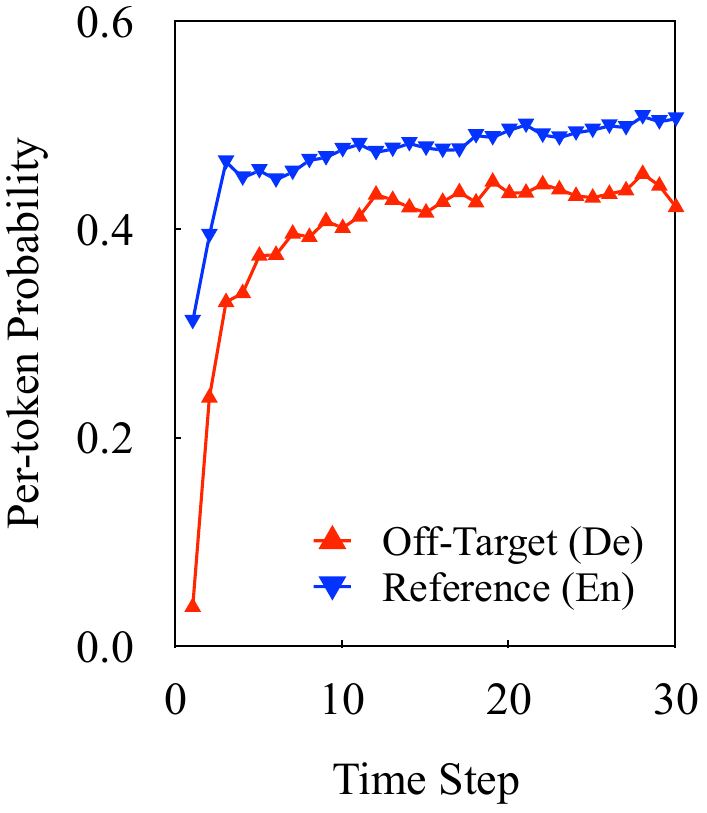}}
    \hfill
    \subfloat[Zero-Shot Fr-De]{
    \includegraphics[height=0.3\textwidth]{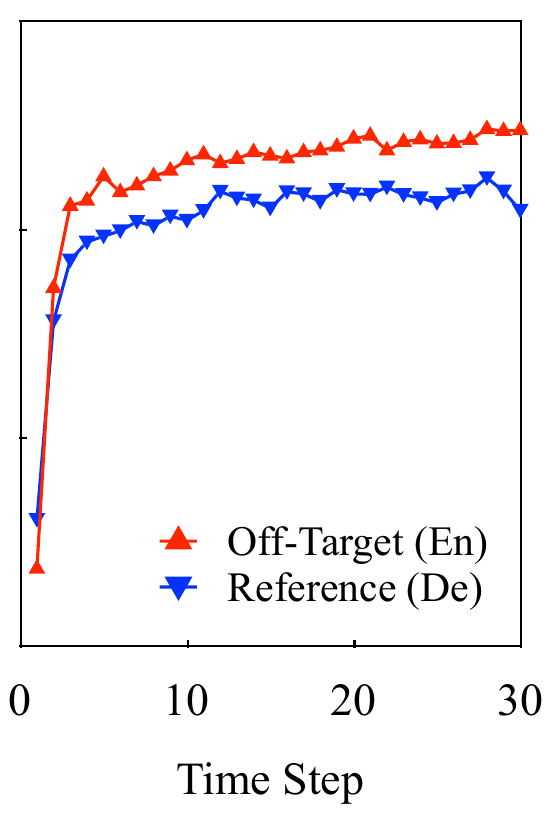}}
    \caption{Per-token probabilities of (a) supervised Fr-En (BLEU$\uparrow$: 38.8; OTR$\downarrow$: 1.6) and (b) zero-shot Fr-De (BLEU$\uparrow$: 5.4; OTR$\downarrow$: 74.9) translations. Higher probabilities are expected for the on-target references (``Reference''), and lower probabilities are expected for the off-target distractor translations (``Off-Target'').}
    \label{fig:per-token-prob-vanilla}
\end{figure}

\subsection{\bf\em Uncertain Prediction Causes Off-Target Issues}

To investigate how MNMT models generate off-target translations, we follow~\cite{Ott2018ICML} to analyze the model confidence in the target language. Specifically, we compute the average probability at each time step across a set of sentence pairs.
In addition to the ground-truth reference sentence, we also consider a ``distractor'' translation in the off-target language for each source sentence.
Figure~\ref{fig:per-token-prob-vanilla} plots the model confidence for both references (``Reference'') and distractors ( ``Off-Target'') on supervised Fr-En and zero-shot Fr-De tasks. {We find that 94.7\% of the off-target translations in the zero-shot Fr-De task are in English. Therefore, we only present the English off-target translation for simplicity.}
One thing that needs to be mentioned is when calculating the per-token probability of En/De, we do not need to consider which tokens belong to English, German, or both. For each French sentence, we have one English reference sentence and one German reference sentence. Given the same French sentences as the input, we calculate the average per-token probability of generating the English/German reference sentences.

From Figure~\ref{fig:per-token-prob-vanilla} we can find that, different from the supervised translation, the zero-shot translation shows a surprisingly higher confidence in the off-target distractors. Accordingly, the model tends to generate more off-target translations (i.e., 74.9\% vs. 1.6\%).

\vspace{5pt}
In uncertainty theory\cite{Steck1989ExperimentationAU}, the uncertainties for an engineering system can come from model design,  named intrinsic uncertainties, or from model calibrations using uncertain data, named extrinsic uncertainty. In the following sections, we will connect the uncertain prediction problem to the language uncertainty from both data (Section~\ref{sec:data-uncertainty}) and model (Section~\ref{sec:model-uncertainty}). Based on these findings, we provide simple and effective solutions to mitigate the data and model uncertainty.

\section{Extrinsic Data Uncertainty} 
\label{sec:data-uncertainty}

\begin{table}[t!]
\centering
\caption{Ratios(\%) of off-target noises in the training datasets}
\begin{tabular}{c cccccc}
\toprule
{\bf Training} & \multicolumn{6}{c}{\bf Language Paris (En-)}\\
\cmidrule(lr){2-7}
\bf Data &  \bf Zh & \bf Ja & \bf De & \bf Fr & \bf Ro & \bf Ave.\\
\midrule
OPUS-6 & 1.3 & 1.1 & 8.5 & 9.0 & 9.2 & 5.8\\
WMT-6 & 0.1 & 0.6 & 2.5 & 2.3 & 2.1 & 1.5\\
\bottomrule
\end{tabular}
\label{tab:training-set-off-target}
\end{table}

\subsection{Problem: Data Uncertainty}
The uncertainty in multilingual training data is an important reason for the uncertain prediction in zero-shot translation. As a data-driven approach, MNMT models learn the mappings between languages from the parallel data, and we assume that both the source and target sentences are in the correct languages. However, we find that a portion of training data contains off-target translations, mainly in English. We guess it is because the dataset is collected by translated corpus mainly from English, such as the TED subscripts translated from English. For some sentences, the translators don’t know how or don’t think they need to be translated, so they leave the text in English, leading to the English-English pairs.
Table~\ref{tab:training-set-off-target} lists the statistics, where we observe a high off-target ratio in both OPUS-6 (i.e., 5.8\%) and WMT-6 (i.e., 1.5\%).
A previous study on bilingual MT~\cite{Ott2018ICML} suggests that 1\% to 2\% of such data noises can make the NMT model highly uncertain and tend to produce translations in the source language.
We believe that similar uncertainty issues will also occur in MNMT models, especially for zero-shot translation where no supervision signal (from parallel data) exists.

\subsection{Solution: Data Denoising}
We utilize data denoising to make sure the MNMT model learns a more correct mapping between languages from the training data.
Specifically, we adopt the \texttt{langid} tool to identify the off-target sentence pairs in the training data of each parallel data and remove them to build a clean dataset. The clean dataset is then used for training the MNMT models.
Without the distraction from the off-target sentence pairs, the MNMT model is expected to be more confident in the target languages. As a result, we can reduce the off-target ratio and improve the performance of zero-shot translation.

Table~\ref{tab:data-denoising} lists the results of removing off-target noises for both OPUS-6 and WMT-6 datasets. The data denoising method significantly improves the zero-shot translation performance by greatly reducing the off-target issues. However, there are still around 10\% off-target translations unsolved, which we attribute to the intrinsic model uncertainty due to the nature of multilingual learning (\S~\ref{sec:model-uncertainty}).

\begin{table}[t]
\centering
\begin{tabular}{c rr rr}
\toprule
\bf Training &  \multicolumn{2}{c}{\bf Supervised}  & \multicolumn{2}{c}{\bf Zero-Shot}\\
\cmidrule(lr){2-3}\cmidrule(lr){4-5}
\bf Data    &   \bf BLEU$\uparrow$  & \bf OTR$\downarrow$    &   \bf BLEU$\uparrow$  & \bf OTR$\downarrow$\\
\midrule
\multicolumn{5}{c}{\bf OPUS-6 Data}\\
Raw Data   &  27.2 & 1.9 & 10.2 & 32.1\\
~~~ + Denoise &   27.1 & 1.5 & 14.0 & 10.0\\
\midrule
\multicolumn{5}{c}{\bf WMT-6 Data}\\
Raw Data    &  28.8 & 1.7 & 13.3 & 22.5\\
~~~ + Denoise &  28.8 & 1.6 & 15.3 & 10.4\\
\bottomrule
\end{tabular}
\caption{Results of data denoising for \textsc{T-Enc} model.}
\label{tab:data-denoising}
\end{table}

\section{Intrinsic Model Uncertainty}
\label{sec:model-uncertainty}

\subsection{Problem: Over-Estimation on Off-Target Vocabulary}

The uncertainty inside the MNMT model is another reason for the uncertain prediction in zero-shot translation. To enhance the knowledge transfer between languages, researchers seek to train a single model with parameters shared by different languages, including the vocabulary and the corresponding embedding matrix. 
However, the shared vocabulary also introduces uncertainty to the decoder output. Theoretically, the MNMT model is allowed to predict any token in the vocabulary (e.g., a word in the source language can correspond to multiple translations in different forms and languages~\cite{jiao2021self}), preserving the possibility of decoding into a wrong language.
Such language uncertainty can be avoided with the supervision of parallel data, which is unavailable for zero-shot translation.

\begin{figure}[t]
    \centering 
    \includegraphics[height=0.28\textwidth]{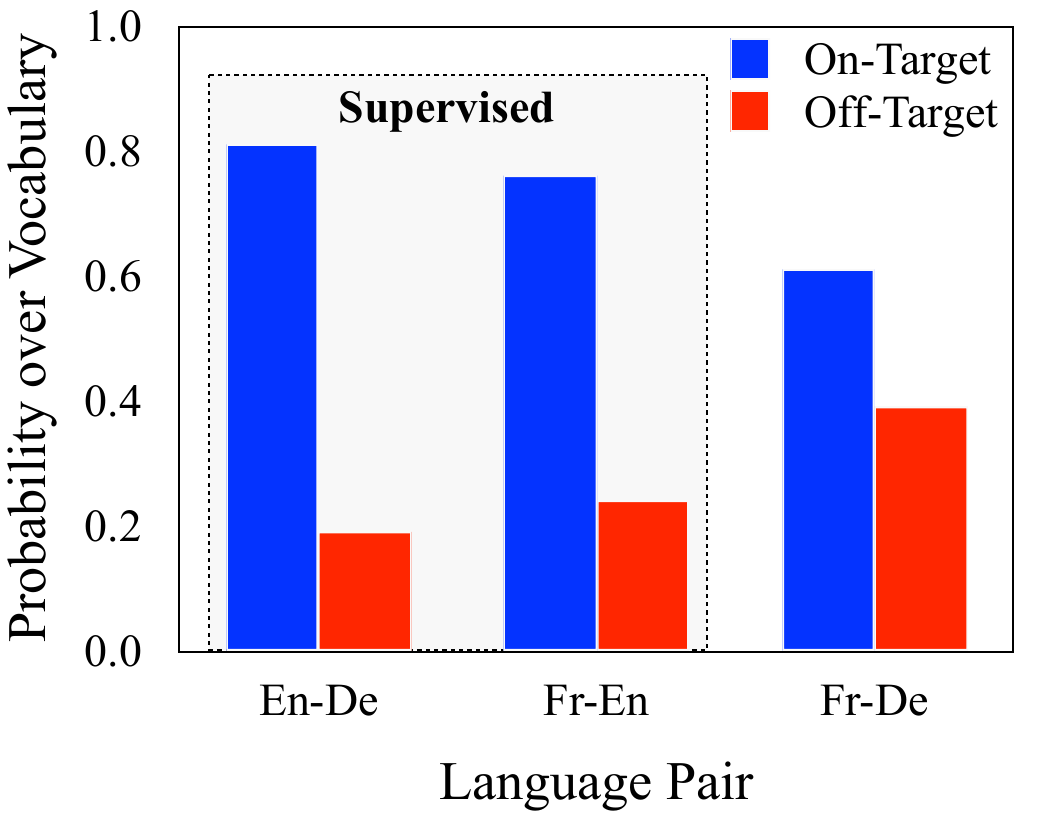}
    \caption{Probability over the vocabulary of supervised (En-De, Fr-En) and zero-shot Fr-De translations. MNMT model over-estimates the off-target vocabulary for zero-shot translation.}
    \label{fig:prob-over-vocab}
\end{figure}

Empirically, we compute the prediction distribution over the whole vocabulary for each token in the reference sentences. Then, we calculate how much of the probability mass is assigned to the target language (``On-Target'') based on its individual vocabulary, and how much to the others (``Off-Target''). 
Figure~\ref{fig:prob-over-vocab} plots the results on the zero-shot Fr-De translation. For reference, we also plot the related supervised En-De and Fr-En translations.
Obviously, for supervised translation, the vast majority of the probability mass is assigned to the target language. However, for zero-shot translation, more probability mass (i.e., around 39\%) is assigned to off-target languages, thus leading to serious off-target issues.

\subsection{Solution}

Based on the above findings, we propose two methods to reduce over-confidence in off-target vocabulary, which differ in whether to use the off-target vocabulary in training.

\vspace{3pt}
\subsubsection{\bf \em Vocabulary Masking}
One straightforward solution to model uncertainty is to constrain the probability distributions only on the vocabulary of the target language by masking the output logits on the off-target vocabulary. \cite{Brard2021EfficientIF} proposed a target vocabulary filtering method that filters the model’s target vocabulary and embeddings to only contain the token from target languages, aiming to accelerate the inference times. Similar to their method, we filter the tokens from other languages to mitigate the off-target issues.
Specifically, we extract a language-specific vocabulary $V_l$ for each language $l\in\mathcal{L}$ from the full vocabulary $V$ ($V_l \subset V$). We first build a BPE vocabulary shared by all languages, which is the same one used for the vanilla MNMT model. We then construct the language-specific vocabulary by counting the BPE tokens in the segmented training data of the corresponding language. Note that different language-specific vocabularies can have shared tokens. 
For example, the English-specific vocabulary shares 33\% tokens with German-specific vocabulary on the OPUS-6 data (see Table~\ref{tab:language-specific-vocabulary} for more details).

This method can be applied in both training and inference. When predicting target tokens, we mask the tokens that do not appear in the vocabulary $V_T$ of target language $T$. Formally, the output probability of the token $y$ is calculated as:
\begin{align}
P_{\theta}(y|{\bf h}_t)=\left\{
\begin{aligned}
& \frac{\exp({\bf h}_t^\top {\bf w}_y)}{\sum_{y' \in V_T}\exp({\bf h}_t^\top {\bf w}_{y'})},         &&  y\in V_T \\
& 0,  && \mathrm{otherwise} \nonumber \\
\end{aligned}
\right.
\end{align}
where 
${\bf h}_t$ is the hidden state at time step $t$, and 
${\bf w}_y$ is the word embedding of the token $y$ in~\cite{Vaswani2017AttentionIA, press2016using}.

\vspace{3pt}
\subsubsection{\bf \em Unlikelihood Training}

While the \textit{vocabulary masking} method can successfully reduce the probabilities of translations in the wrong languages, the performance may be limited by two factors: (1) The language-specific vocabularies need to be carefully partitioned for different languages, especially for similar ones (e.g., English and German). (2) The isolation of vocabularies may hinder knowledge transfer across languages. To avoid these limitations, we incorporate the \textit{unlikelihood training objective}~\cite{welleck2019neural} for MNMT, which forces the model to assign lower probabilities to unlikely generations.

Formally, the original likelihood training loss on a translation sentence pair is expressed as:
\begin{align}
    \mathcal{L}_{\rm Likelihood} = -\sum_{t=1}^{T} \log P_{\theta}(y|{\bf x}, {\bf y}_{<t}^{l_c}), \nonumber
\end{align}
where $l_c$ denotes the correct language tag for the target sentence $\mathbf{y}$. This training loss encourages the model to generate {\em on-target translation}. 

We design an additional unlikelihood loss to penalize the {\em off-target translation}.
To simulate the off-target translation, for each sentence pair we change the target language tag to another wrong language $l_w$ and form the negative candidate. Then the unlikelihood training loss is defined as:
\begin{align}
    \mathcal{L}_{\rm Unlikelihood} &= -\sum_{t=1}^{T} \log (1 - P_{\theta}(y|{\bf x}, {\bf y}_{<t}^{l_w})). \nonumber
\end{align}
The final loss is the combination of the above two:
\begin{align}
    \mathcal{L} &= \mathcal{L}_{\rm Likelihood} + \alpha \mathcal{L}_{\rm Unlikelihood}. \nonumber
\end{align}
In this way, we provide supervision for zero-shot directions by penalizing the off-target translations (i.e., the mismatch between the target language tag and the target sentence).
We follow Welleck et al~\cite{welleck2019neural} to fine-tune the pre-trained MNMT model with the combined loss for $K$ steps.

\subsection{Ablation Study}

\subsubsection{ Ablation Study on Vocabulary Masking}

\begin{table}[t!]
    \centering
    \caption{Statistics of language-specific vocabulary used in vocabulary masking on OPUS-6 data. Each number denotes the number of the shared tokens between the vocabulary of the languages in rows and columns.}
    \begin{tabular}{c rrrrrr}
    \toprule
        &   \bf En  &   \bf De  &   \bf Fr  &   \bf Ja  &   \bf Ro  &   \bf Zh\\
    \midrule
    \bf En  &   \bf 17.2K   \\
    \bf De  &   5.7K    &   \bf 16.2K   \\
    \bf Fr  &   6.0K    &   5.5K    &   \bf 14.9K   \\
    \bf Ja  &   0.9K    &   0.8K    &   0.7K    &   \bf 9.0K\\
    \bf Ro  &   3.8K    &   4.2K    &   5.0K    &   0.3K    &   \bf 12.1K\\
    \bf Zh  &   2.0K    &   1.2K    &   1.2K    &   2.4K    &   1.2K    &   \bf 15.4K\\
    \bottomrule
    \end{tabular}
    \label{tab:language-specific-vocabulary}
\end{table}

\vspace{3pt}
\paragraph{\bf \em Statistics of Language-Specific Vocabularies}
For the language masking approach, we need to extract a language-specific vocabulary for each language from the full vocabulary, and different language-specific vocabularies can have shared tokens. Table~\ref{tab:language-specific-vocabulary} lists the vocabulary statistics on OPUS-6 data. For example, the size of English vocabulary is 17.2K, which shares 5.7K tokens with the German vocabulary.

\begin{table}[h]
\centering
\caption{Impact of vocabulary masking used in training and/or inference on OPUS-6 data.}
\begin{tabular}{cc rr rr}
\toprule
\multicolumn{2}{c}{\bf Mask in}  & \multicolumn{2}{c}{\bf Supervised}  & \multicolumn{2}{c}{\bf Zero-Shot}\\
\cmidrule(lr){1-2}\cmidrule(lr){3-4}\cmidrule(lr){5-6}
\bf Train   &   \bf Infer.    &   \bf BLEU$\uparrow$  & \bf OTR$\downarrow$    &   \bf BLEU$\uparrow$  & \bf OTR$\downarrow$ \\
\midrule
\multicolumn{6}{c}{\bf \em \textsc{S-Enc-T-Dec} MNMT Models}\\
\texttimes  &   \texttimes  &  27.1 & 1.9 & 12.3 & 20.6  \\
\hdashline
\texttimes  &   \checkmark  & 27.1 & 1.8 & \bf  14.4 & \bf 7.2  \\
\checkmark  &   \checkmark  & 27.1 & 1.8 & 13.9 & 10.5  \\
\midrule
\multicolumn{6}{c}{\bf \em \textsc{T-Enc} MNMT Models}\\
\texttimes  &   \texttimes  &  27.2 & 1.8 & 10.2 & 32.1 \\
\hdashline
\texttimes  &   \checkmark  &  27.2 & 1.8 & \bf 13.1 & \bf 12.7 \\
\checkmark  &   \checkmark  &  27.2 & 1.8 & 12.5 & 18.6 \\
\bottomrule
\end{tabular}
\label{tab:ablation-vocab-masking}
\end{table}

\vspace{3pt}
\paragraph{\bf \em Language Masking in Training or Inference}

As aforementioned, the proposed vocabulary masking method can be used in both training and inference. Table~\ref{tab:ablation-vocab-masking} presents the results of different masking strategies.
Applying vocabulary masking to the vanilla MNMT model during inference significantly improves zero-shot translation performance by remedying off-target issues, which demonstrates the effectiveness of vocabulary masking. However, further including vocabulary masking into the training process makes the improvement of zero-shot translation less significant.
One possible reason is that isolating the vocabularies between languages during training 
may hinder cross-lingual knowledge transfer.

\subsubsection{Ablation Study on Unlikelihood Training} 

\begin{figure}[t]
    \centering 
    \subfloat[BLEU$\uparrow$]{
    \includegraphics[height=0.27\textwidth]{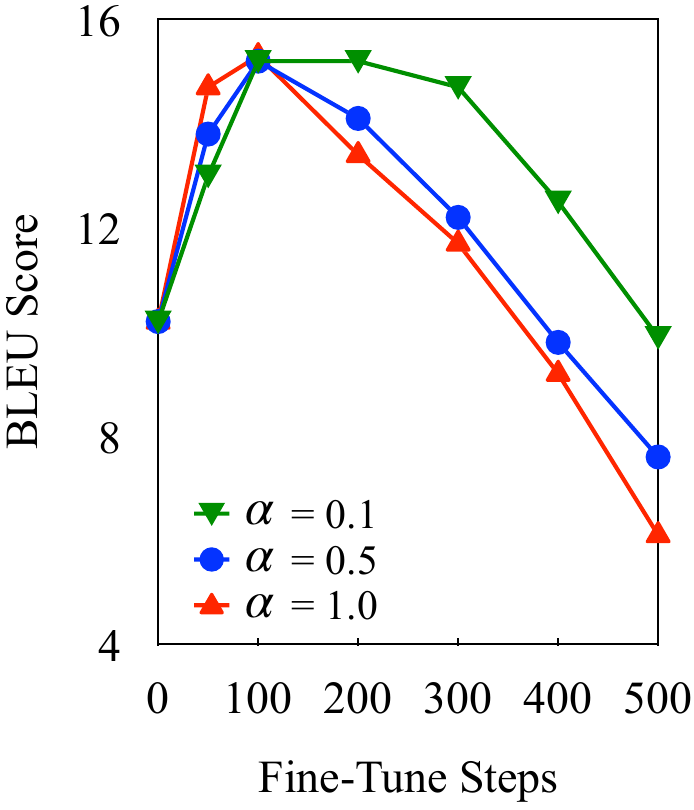}} \hfill
    \subfloat[Off-Target Ratio$\downarrow$]{
    \includegraphics[height=0.27\textwidth]{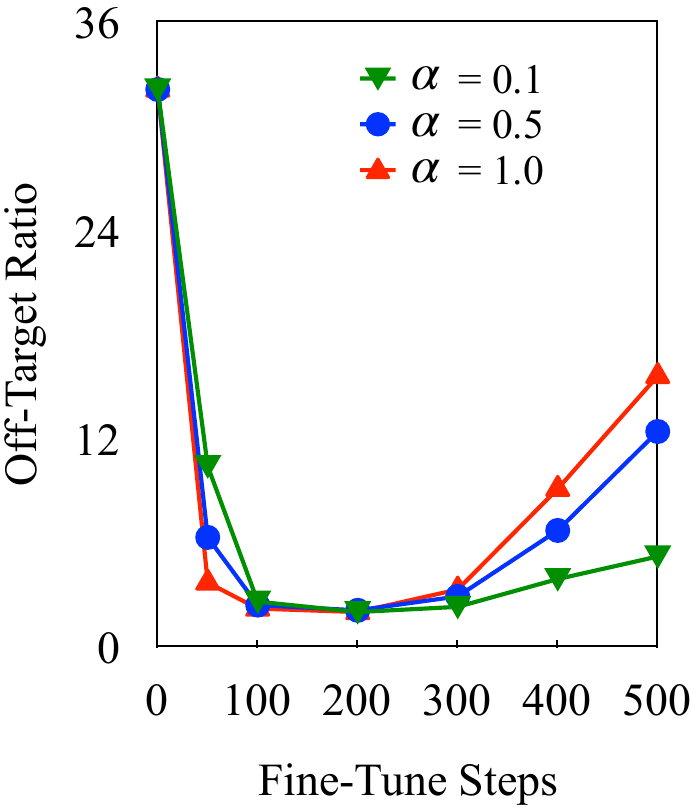}}
    \caption{Impact of interpolation weight $\alpha$ and fine-tune step $K$ on zero-shot translations.} 
    \label{fig:unlikihoood-alpha}
\end{figure}

Figure~\ref{fig:unlikihoood-alpha} shows the impact of the interpolation weight $\alpha$ and fine-tune steps $K$ on unlikelihood training. The zero-shot performance goes up with the increase of fine-tune steps $K$ until $K=100$ for all interpolation weights, and declines when fine-tuning for more steps. One possible reason is that the negative examples are semantically equivalent sentence pairs, while the target language tag is replaced with a wrong tag beyond the target language. The mismatch between the target language tag and the target sentence is a simple pattern, which can be easily learned by the model with as few as 100 steps. Fine-tuning with unlikelihood loss of higher interpolation weights or for more steps potentially harms the cross-lingual transfer ability among semantically equivalent sentences. 
In the following experiments, we use $\alpha=0.1$ and $K=100$ as default for its robust performance.

\begin{table}[t!]
\centering
\caption{Results of mitigating model uncertainty for \textsc{T-Enc} model on raw data without denoising.}
\begin{tabular}{l rr rr}
\toprule
\multirow{2}{*}{\bf Model} &  \multicolumn{2}{c}{\bf Supervised}  & \multicolumn{2}{c}{\bf Zero-Shot}\\
\cmidrule(lr){2-3}\cmidrule(lr){4-5}
    &   \bf BLEU$\uparrow$  & \bf OTR$\downarrow$    &   \bf BLEU$\uparrow$  & \bf OTR$\downarrow$\\
\midrule
\multicolumn{5}{c}{\bf OPUS-6 Data}\\
Vanilla   &  27.2 & 1.9 & 10.2 & 32.1\\
~ + Vocab Mask   & 27.2 & 1.8 & 13.1 & 12.7\\
~ + Unlike Train & 27.2 & 1.5 & 15.2 & 2.2\\
\midrule
\multicolumn{5}{c}{\bf WMT-6 Data}\\
Vanilla    &  28.8 & 1.7 & 13.3 & 22.5\\
~ + Vocab Mask &  28.8 & 1.6 & 14.9 & 10.7\\
~ + Unlike Train & 28.8 & 1.6 & 16.3 & 5.6\\
\bottomrule
\end{tabular}
\label{tab:model}
\end{table}

\subsubsection{Comparison Results} Table~\ref{tab:model} lists the results of vocabulary masking and unlikelihood training. Clearly, unlikelihood training consistently outperforms vocabulary masking on zero-shot translation in all cases. We attribute the superiority of unlikelihood training to directly penalizing simulated off-target translation during model training. In the following experiments, we use unlikelihood training to mitigate model uncertainty as default.

\section{Overall Empirical Assessment}
\label{sec:main}

In this section, we investigate whether our approaches can alleviate the off-target issue in zero-shot translation.

\subsection{Translation Performance}

\begin{table*}[t]
\setlength{\tabcolsep}{5pt}
\centering
\caption{BLEU scores and off-target ratios (OTR) of multilingual translation models on the TED58 {\bf supervised} (i.e., 10 English-centric language pairs) and {\bf zero-shot} (i.e., 20 non-English-centric pairs) test sets. Our approaches consistently improve zero-shot translation performance without sacrificing the quality of supervised translation.}
\begin{tabular}{l rr rr rr rr rr rr}
\toprule
\multirow{3}{*}{\bf Model}  &   \multicolumn{4}{c}{\bf OPUS-6 Data}  & \multicolumn{4}{c}{\bf WMT-6 Data} & \multicolumn{4}{c}{\bf OPUS-100 Data}\\
\cmidrule(lr){2-5} \cmidrule(lr){6-9} \cmidrule(lr){10-13}
& \multicolumn{2}{c}{\bf Supervised}  & \multicolumn{2}{c}{\bf Zero-Shot} &  \multicolumn{2}{c}{\bf Supervised}  & \multicolumn{2}{c}{\bf Zero-Shot} &  \multicolumn{2}{c}{\bf Supervised}  & \multicolumn{2}{c}{\bf Zero-Shot}\\
\cmidrule(lr){2-3}\cmidrule(lr){4-5}\cmidrule(lr){6-7}\cmidrule(lr){8-9} \cmidrule(lr){10-11} \cmidrule(lr){12-13}
    &   \bf BLEU$\uparrow$  & \bf OTR$\downarrow$    &   \bf BLEU$\uparrow$  & \bf OTR$\downarrow$    &   \bf BLEU$\uparrow$  & \bf OTR$\downarrow$    &   \bf BLEU$\uparrow$  & \bf OTR$\downarrow$  &   \bf BLEU$\uparrow$  & \bf OTR$\downarrow$    &   \bf BLEU$\uparrow$  & \bf OTR$\downarrow$\\
\midrule
\multicolumn{13}{c}{\bf \em \textsc{S-Enc-T-Dec} MNMT Models}\\
{Vanilla}       & 27.1 & 1.9 & 12.3 & 20.6 & 28.0 & 1.8 & 10.6 & 37.8 & 26.9 & 1.8 & 1.2 & 92.7\\
\hdashline
{Data Denoise}  & 27.2 & 1.5 & 14.1 & 7.0 & 28.0 & 1.7 & 11.1 & 23.9 & 27.0 & 1.5 & 4.8 & 45.0\\
{Vocab Mask}    & 27.1 & 1.8 & 14.4 & 7.2 & 27.8 & 1.7& 14.0 & 17.3 & 26.9 & 1.8 & 9.4 & 23.1\\
{~~~~+Data Denoise}          & 27.2 & 1.5 & 14.8 & 3.1 & 27.9 & 1.7 & 14.2  &  13.4 & 27.0 & 1.5 & 10.0 & 16.4 \\
{Unlikelihood Training}  & 27.1 & 1.7 &  15.3 & 1.7 & 28.0 & 1.7 & 16.4 & 4.0 & 27.0 & 1.5 & 12.5 & 2.3 \\
{~~~~+Data Denoise}   & 27.2 & 1.5 & \bf 15.6 & \bf  1.1 & 28.0 & 1.8 & \bf 17.4 &\bf  2.4 & 27.0 & 1.5 &\bf  12.6 &\bf  1.6\\
\midrule
\multicolumn{13}{c}{\bf \em \textsc{T-Enc} MNMT Models}\\
{Vanilla}       & 27.2  & 1.8 & 10.2 & 32.1 & 28.8 & 1.7 & 13.3 & 22.5 & 26.9 & 1.8 & 7.5 & 38.4  \\
\hdashline
{Data Denoise}  & 27.1 & 1.5  & 14.0 & 10.0 & 28.8 & 1.6 & 15.3 & 10.4 & 26.9 & 1.6 &8.8 & 21.4 \\
{Vocab Mask}    & 27.2 & 1.8 & 13.1 & 12.7  & 28.8 & 1.6 & 14.9 & 10.7 & 26.9 & 1.8 & 10.1 & 15.2 \\
{~~~+Data Denoise}          & 27.2 & 1.5 & 14.4 &  5.1  & 28.8 & 1.6 &  15.7 &  6.1 & 26.9 & 1.6 & 11.2 & 10.9\\
Unlikelihood Training & 27.1 & 1.5 & 15.0 & 2.6 & 28.8 & 1.6 & 16.3 & 5.9 & 26.9 & 1.7 & 12.6 & 5.7\\
{~~~~+Data Denoise}    &   27.2 & 1.5 &\bf   15.2 &\bf   2.2 & 28.8 & 1.6 &\bf   16.8 &\bf   4.2 & 26.9 & 1.7 &\bf   13.1 &\bf   3.5\\ 
\bottomrule
\end{tabular}
\label{tab:main-bleu-otr}
\end{table*}

Table~\ref{tab:main-bleu-otr} lists the results of both supervised and zero-shot translations. Clearly, both data denoising, vocabulary masking, and unlikelihood training can significantly improve the zero-shot translation performance in all cases. 
Combing them together achieves the best performance, demonstrating the complementarity between data uncertainty and model uncertainty. We also demonstrate the effectiveness of our proposed method using different metrics like COMET and ChrF, as shown in Table~\ref{tab:main-comet-chrf}.

\begin{table}[t]
\centering
\caption{Results of MNMT models trained on the OPUS-6 dataset measured by other evaluation metrics.}
\begin{tabular}{l rr rr}
\toprule
\multirow{2}{*}{\bf Model}  
& \multicolumn{2}{c}{\bf Supervised}  & \multicolumn{2}{c}{\bf Zero-Shot}\\
\cmidrule(lr){2-3}\cmidrule(lr){4-5}
    &   \bf COMET$\uparrow$  & \bf ChrF$\uparrow$    &   \bf COMET$\uparrow$ & \bf ChrF$\uparrow$   \\
\midrule
\multicolumn{5}{c}{\bf \em \textsc{S-Enc-T-Dec} MNMT Models}\\
Vanilla & 0.168  & 51.0 &-0.297 & 26.0\\
+ Data Denoise & 0.169 & 51.4  &-0.193 & 29.1\\
\hdashline
~~ + Vocab Mask & 0.169 & 51.4 & -0.125 &30.0\\
~~ + Unlike Train &  0.169 & 51.4 &-0.098 & 30.8\\
\midrule
\multicolumn{5}{c}{\bf \em \textsc{T-Enc} MNMT Models}\\
Vanilla         & 0.316 &  46.4 & -0.336 & 22.9 \\
+ Data Denoise  & 0.319 & 46.4  & -0.187 & 28.8\\
\hdashline
~~ + Vocab Mask & 0.319 & 46.4  & -0.142 & 29.3\\
~~ + Unlike Train & 0.320 & 46.4 & -0.104 & 30.8\\
\bottomrule
\end{tabular}
\label{tab:main-comet-chrf}
\end{table}

\paragraph{Larger-Scale Imbalanced Datasets} 
In addition to the small-scale balanced OPUS-6 data, we also validate our approaches on the larger-scale imbalanced datasets (i.e. 111.8M WMT-6 data with 6 languages and 55.0M OPUS-100 with 100 languages). Generally, the off-target issues are more severe in imbalanced scenarios. For example, the zero-shot translation almost crashes on imbalanced OPUS-100 data with 92.7\% of off-target translation. Our approaches perform well by reducing the off-target issues to as low as 1.6\% to 2.4\%, which are close to that on the small-scale balanced data (i.e. 1.1\% on OPUS-6).
These results demonstrate the scalability of our approaches to massively multilingual translation tasks.

\paragraph{Different Tagging Strategies}
There are considerable differences between \textsc{T-Enc} and \textsc{S-Enc-T-Dec} models, which differ in how to attach the language tags. \textsc{T-Enc} performs significantly better on imbalanced datasets (especially on OPUS-100), while performs worse on balanced OPUS-6 data than its \textsc{S-Enc-T-Dec} counterpart. Our approaches can consistently improve zero-shot performance on top of \textsc{T-Enc} in all cases, demonstrating the universality of the proposed approaches. S-Enc-S-Dec is a not widely adopted tagging manner, which encodes both source and target language in the encoder, due to its worse zero-shot translation performance~\cite{Wu2021LanguageTM}. In Appendix Table ~\ref{tab:senc-tenc}, we also show that our proposed methods can improve the zero-shot translation performance and mitigate the off-target issues on top of the S-Enc-S-Dec model.

With the help of our approaches, \textsc{S-Enc-T-Dec} produces better overall performance than \textsc{T-Enc}. One possible reason is that \textsc{S-Enc-T-Dec} is better at modeling language mapping by explicitly identifying the source and target languages. Meanwhile, the side-effect of over-fitting on supervised mapping can be almost solved by our approaches.

\subsection{Prediction Uncertainty}

In this section, we present a qualitative analysis to provide insights into where our approaches reduce off-target translations. 

\begin{figure}[t]
    \centering
    \includegraphics[height=0.28\textwidth]{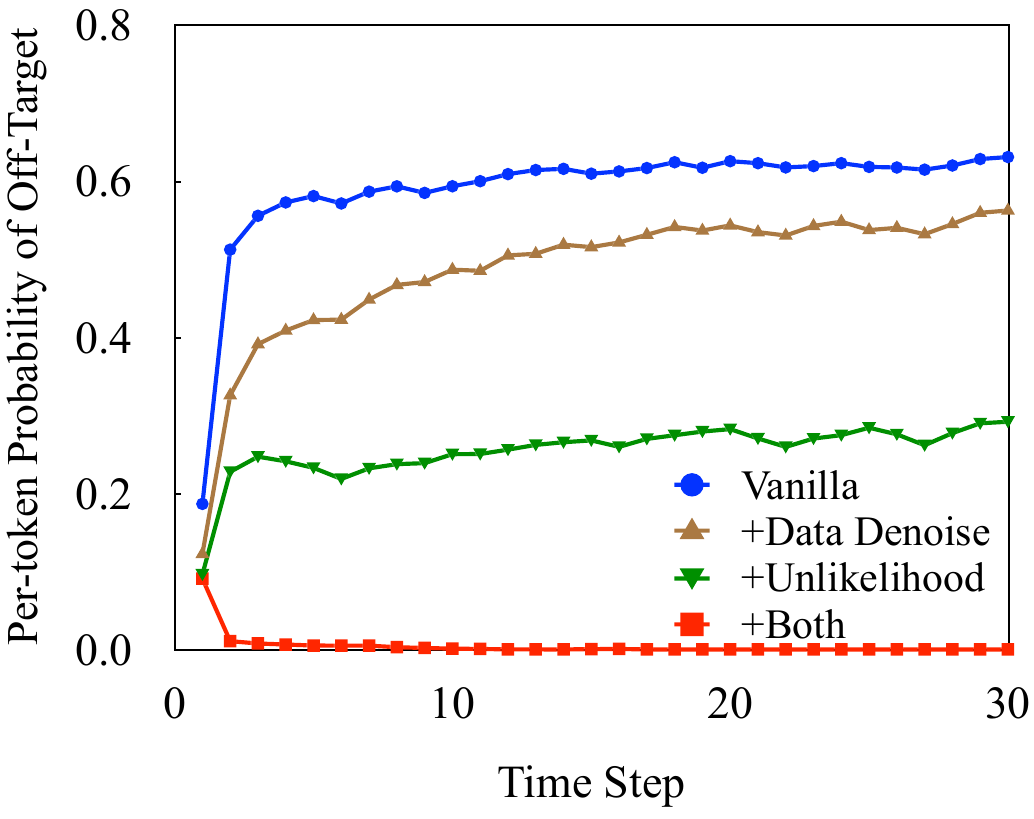}
    \caption{Per-token probabilities of off-target test sentences for zero-shot Fr-De translations for \textsc{T-Enc} model with our methods on OPUS-6 data.}
    \label{fig:per-token-prob-ours}
\end{figure}

Figure~\ref{fig:per-token-prob-ours} shows the prediction probabilities on the off-target translation that are produced by the vanilla \textsc{T-Enc} model on zero-shot translation. 
Clearly, data denoising and unlikelihood training consistently reduce the model confidence on the off-target translation, which reconfirms our claim that extrinsic data uncertainty and intrinsic model uncertainty are responsible for the uncertain prediction of target languages.
Specifically, we find that data denoising reduces the confidence on the first few tokens of off-target translations noticeably, 
while unlikelihood training consistently reduces all the tokens. 
Combining them together (``+Both'') can surprisingly reduce the per-token probability of off-target translation to zero. 
The reason is that likelihood training on these off-target noises could encourage the model to generate off-target translation, which partially counteracts the effect of unlikelihood training that prevents the model from generating off-target translation.
Therefore, denoising such off-target noises can further improve the performance of unlikelihood training.

\subsection{Comparison with Previous Work}

We compare our methods with three recent works on improving zero-shot translation: (1) RemoveRes.~\cite{Liu2021ImprovingZT} that removes residual connections in an encoder layer to disentangle the positional information; (2) AE Loss~\cite{Wang:2021:EMNLP} that introduces a denoising autoencoding loss to implicitly maximize the probability distributions for zero-shot directions; (3) Contrastive loss~\cite{Pan2021ContrastiveLF} that maximize the cosine similarly between the encoder representation of sentences of different languages. We reimplement these methods and compare the translation performance. 
Table~\ref{tab:comparison} lists the results on OPUS-6 data, which shows that our methods can consistently outperform their methods. 
The improvement is much larger on the noisy raw data, which we attribute to the advantage of our approach in directly penalizing off-target translations.

\begin{table}[t]
\centering
\caption{Comparison with previous work on improving zero-shot translation. ``Clean Data'' denotes filtering off-target noises using our data denoising method.} 
\begin{tabular}{l rr rr}
\toprule
\multirow{2}{*}{\bf Model} &  \multicolumn{2}{c}{\bf Supervised}  & \multicolumn{2}{c}{\bf Zero-Shot}\\
\cmidrule(lr){2-3}\cmidrule(lr){4-5}
    &   \bf BLEU$\uparrow$  & \bf OTR$\downarrow$    &   \bf BLEU$\uparrow$  & \bf OTR$\downarrow$\\
\midrule
\multicolumn{5}{c}{{\bf Raw Data}}\\
Vanilla &   27.2 & 1.8 & 10.2 & 32.1\\
~~+ RemoveRes. & 26.7 & 1.8 & 12.8 & 21.7 \\
~~+ AE Loss  & 26.7 & 2.0 & 12.2  & 21.0\\
~~+ Contrastive Loss  & 27.0 & 1.8 & 13.5 & 16.4 \\
~~+ Ours & 27.1 & 1.5 & 15.0 & 2.6\\
\midrule
\multicolumn{5}{c}{{\bf Clean Data ({\em with data denoising})}}\\
Vanilla &  27.1 & 1.5 &  14.0 & 10.0\\
~~+ RemoveRes. & 27.1 &1.5 & 14.2 & 5.6\\
~~+ AE Loss  & 26.3 & 1.5 & 14.5 & 5.1 \\
~~+ Contrastive Loss  & 27.1 & 1.5 & 14.8 & 3.9 \\
~~+ Ours & 27.2 & 1.5 & 15.2 & 2.2\\
\bottomrule
\end{tabular}
\label{tab:comparison}
\end{table}

\section{Related Work}
\label{sec-related}

\subsection{Improving Zero-Shot Translation}

A number of recent efforts have explored ways to improve zero-shot translation by mitigating the off-target issue. One thread of work focuses on modifying the model architecture. Zhang, et al.~\cite{Zhang2020ACL} added a target language-aware linear transformation between the encoder and the decoder to enhance the freedom of multilingual NMT in expressing flexible translation relationships. Liu, et al. ~\cite{Liu2021ImprovingZT} found that the off-target issue can be alleviated by removing residual connections in an encoder layer. And Wu~\cite{Wu2021LanguageTM} showed that adding language tags properly to the model can enhance the consistency of semantic representations and alleviate the off-target issue. Compared with these methods, our work is orthogonal to them and mitigates the off-target issue from an uncertainty perspective without the need to modify the architecture.

Another thread of work introduces auxiliary tasks with additional training losses to help the model training. For example, Al-Shedivat et al.~\cite{AlShedivat2019ConsistencyBA} proposed an agreement-based likelihood training objective that encourages the model to produce equivalent translations of parallel sentences in auxiliary languages. Yang et al.~\cite{Yang2021ImprovingMT} leveraged an auxiliary target language prediction task to regularize decoder outputs to retain information about the target language. And Wang et al.~\cite{Wang:2021:EMNLP} introduced an additional denoising autoencoder objective into the traditional training objective to improve the translation accuracy on zero-shot directions. Our work
 proposes a novel and lightweight method to directly reduce the off-target translation via unlikelihood training.

Besides, researchers also try to generate synthetic data for zero-shot translation pairs in either an off-line\cite{Gu:2019:ACL} or on-line\cite{Zhang2020ACL} manner. However, such methods require additional computational costs in generating data and model training. Also, adding data pairs in the zero-shot translation direction could hurt the performance of supervised translation, which is known as the curse of representation bottleneck in the multilingual translation field\cite{Zhang2020ACL}.
Our work also tackles the off-target issue from a data perspective, but we remove the off-target noises in the original data rather than leveraging additional data, which is not a common practice in MNMT.

Recently, a concurrence work~\cite{Chen2023OnTO} also proposed a method to manipulate the distribution of the multilingual vocabulary, aiming to reduce the off-target issues in zero-shot translation. However, the motivation and method are different from this paper.  \cite{Chen2023OnTO} contributes the off-target issues to the closer lexical distance between two languages' vocabulary and tries to enlarge the lexical distance by adding language-specific tokens to the shared tokens. However, according to our experiments, even between two languages that already have large lexical distances, e.g. German and Chinese, we still observe a high off-target ratio. Our paper contributes the off-target issues to the data uncertainty and model uncertainty, and then proposes several methods to mitigate the uncertainties. Our vocabulary masking method, which masks the tokens of other languages when generating a specific language, aims to reduce the probability assigned to the tokens from other languages. Hence, the motivation and specific operation of our paper are different from \cite{Chen2023OnTO}.

\subsection{Uncertainty in NMT}
 
Closely related to our work, Ott et al.~\cite{Ott2018ICML} analyzed the uncertainty in bilingual machine translation, and attributed it to one specific type of data noise -- copies of source sentences. 
In contrast, we analyze the uncertainty in multilingual machine translation, which is a more complicated scenario, for understanding and mitigating the off-target issues. Besides data uncertainty, we also reveal the intrinsic model uncertainty on the output distributions due to the shared vocabulary across multiple languages.

Recently, Chen et al.~\cite{Chen2022FocusOT} proposed Masked Label Smoothing, which masks the soft label probability of source-side words to zero. The operation is similar to our vocabulary masking method. However, they focus on supervised NMT for better calibration and mitigating copy behavior, while we focus on zero-shot translation to mitigate the off-target issues.  In addition, our proposed methods for reducing model uncertainty by either masking out off-target vocabularies or penalizing off-target training examples are carefully designed for the multilingual scenario.

\section{Conclusion}

We present a comprehensive study of the off-target issues in zero-shot translation. We empirically show that the off-target noises in training examples and the shared vocabulary across languages bias MNMT models to over-estimate the translation hypotheses in off-target languages. In response to this problem, we propose several lightweight and complementary approaches to mitigate the uncertainty issues, which can significantly improve zero-shot translation performance with no or only marginal additional computational costs.

In the future, we plan to explore the uncertainty of large MNMT models trained on more complicated datasets~\cite{Fan2021BeyondEM,schwenk-etal-2021-wikimatrix}, as well as that of large language models (LLMs), which also support the zero-shot translation~\cite{jiao2023chatgpt}.

\section*{Acknowledgment}
The work described in this paper was supported by the Research Grants Council of the Hong Kong Special Administrative Region, China (No. CUHK 14206921 of the General Research Fund).

\bibliography{anthology,custom}
\bibliographystyle{IEEEtran}

\section*{Appendix}

\begin{table}[h]
\centering
\caption{BLEU scores and off-target ratios (OTR) of S-Enc-T-Enc Models on TED58 test set}
\begin{tabular}{l rr rr}
\toprule
\multirow{2}{*}{\bf Model} &  \multicolumn{2}{c}{\bf Supervised}  & \multicolumn{2}{c}{\bf Zero-Shot}\\
\cmidrule(lr){2-3}\cmidrule(lr){4-5}
    &   \bf BLEU$\uparrow$  & \bf OTR$\downarrow$    &   \bf BLEU$\uparrow$  & \bf OTR$\downarrow$\\
\midrule
\multicolumn{5}{c}{{\bf OPUS-6 Data}}\\
Vanilla &  27.1 & 2.0 & 8.5 & 43.7\\
+ Data Denoise & 27.1 & 1.8 & 10.3 & 30.9 \\
\hdashline
~~ + Vocab Mask & 27.0 & 1.9 & 12.7 & 18.4\\
~~ + Unlike Train & 27.1 & 1.9 & 14.9 & 3.8 \\
\midrule
\multicolumn{5}{c}{{\bf WMT-6 Data}}\\
Vanilla &  28.6 & 1.7 & 7.8 & 44.2\\
+ Data Denoise & 28.6 & 1.7 & 9.4 & 35.4 \\
\hdashline
~~ + Vocab Mask & 28.6 & 1.7 & 13.9 & 12.1\\
~~ + Unlike Train & 28.6 & 1.7 & 15.3 & 5.0 \\
\bottomrule
\end{tabular}
\label{tab:senc-tenc}
\end{table}

\begin{IEEEbiography}[{\includegraphics[width=1in,height=1.25in,clip,keepaspectratio]{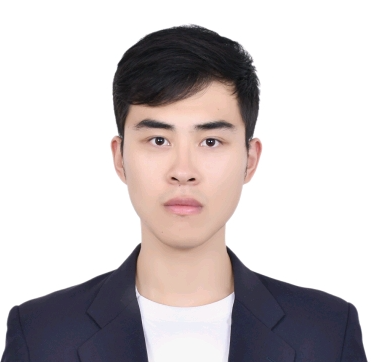}}]{Wenxuan Wang}
is a Ph.D. candidate at the Department of Computer Science and Engineering at The Chinese University of Hong Kong, advised by Prof. Michael R. Lyu. He received his B.S. from Huazhong University of Science and Technology in 2017. His research interests are mainly in the reliability of natural language processing models and software, such as machine translation models and pre-training language models.
\end{IEEEbiography}

\begin{IEEEbiography}[{\includegraphics[width=1in,height=1.25in,clip,keepaspectratio]{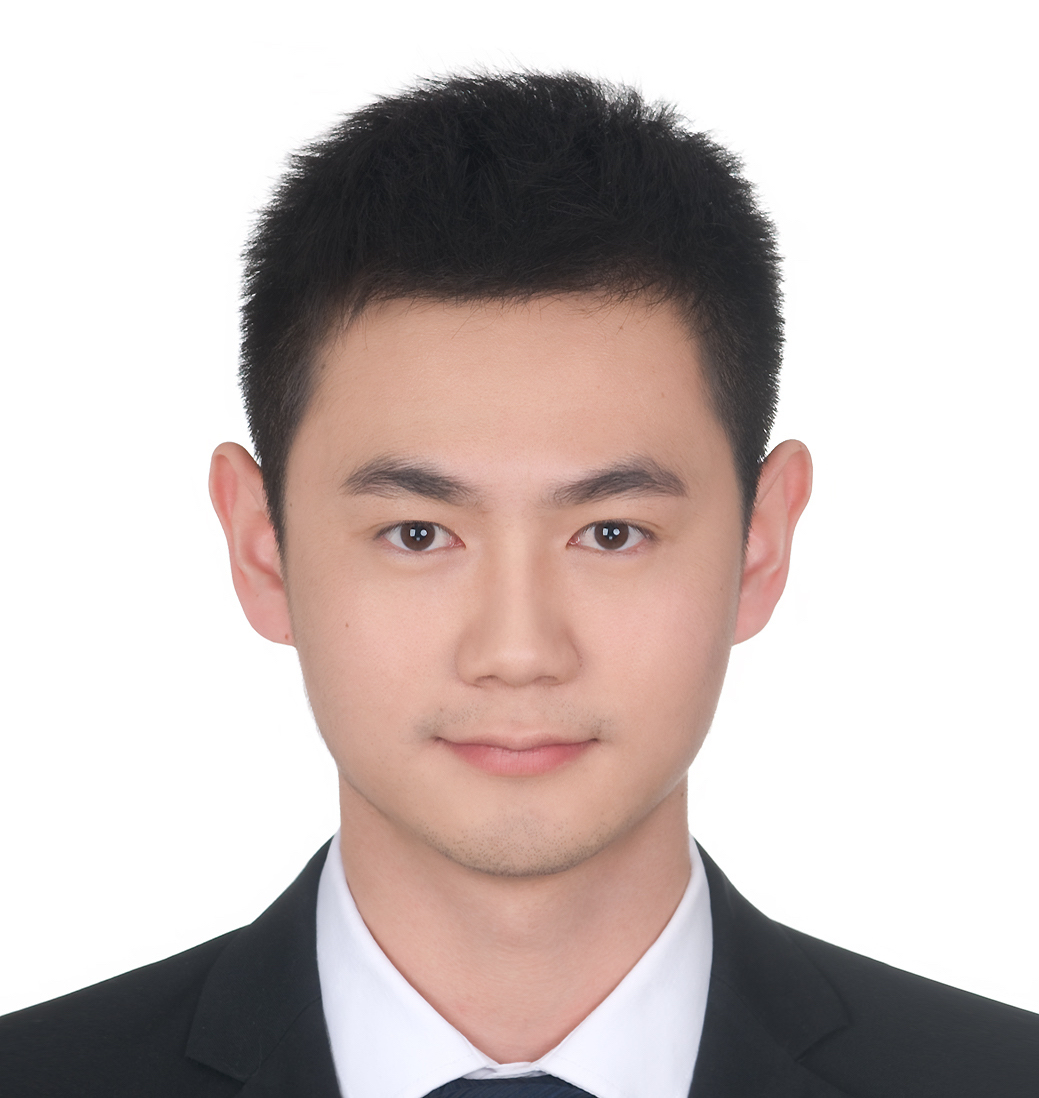}}]{Wenxiang Jiao}
is a senior researcher at the Tencent AI Lab, Shenzhen, China. He received his Ph.D. degree from the Chinese University of Hong Kong in 2021, under the supervision of Prof. Irwin King and Prof. Michael R. Lyu. Before that, he received his Bachelor's degree and MPhil degree at Nanjing University in 2015 and 2017, respectively. His research interests include conversational emotion recognition, neural machine translation, and multilingual pretraining, and has published papers in top-tier conferences and journals such as ACL, EMNLP, NAACL, AAAI, TASLP, etc. He has won the 1st place at WMT~2022 Large-Scale Machine Translation Evaluation for African Languages (Constrained Track), and the Best Paper Award of Multilingual Representation Learning Workshop in EMNLP 2022.
\end{IEEEbiography}

\begin{IEEEbiography}[{\includegraphics[width=1in,height=1.25in,clip,keepaspectratio]{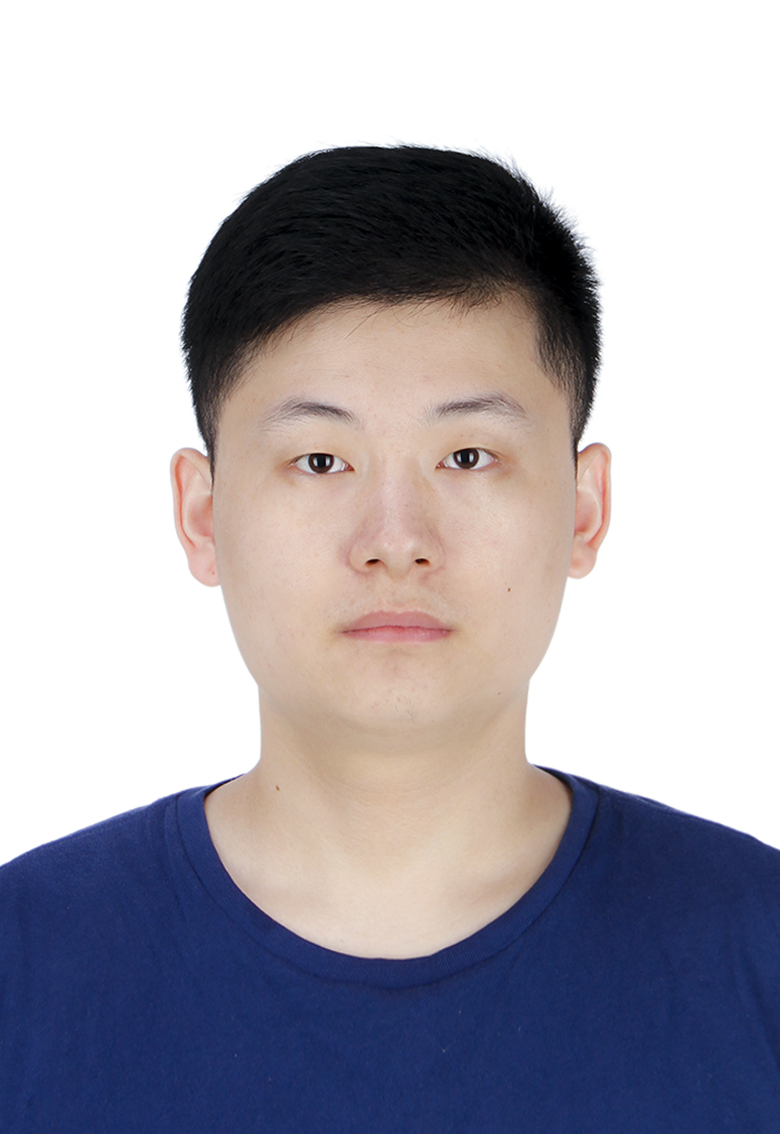}}]{Shuo Wang}
 is a fifth-year Ph.D. candidate at the Department of Computer Science and Technology (Dept. of CST) at Tsinghua University. He is advised by Prof. Yang Liu. Before that, he received his B.S. degree from the Dept. of CST, Tsinghua, in 2018. His research interests include trustworthy neural machine translation and multilingual natural language processing.
\end{IEEEbiography}

\begin{IEEEbiography}[{\includegraphics[width=1in,height=1.25in,clip,keepaspectratio]{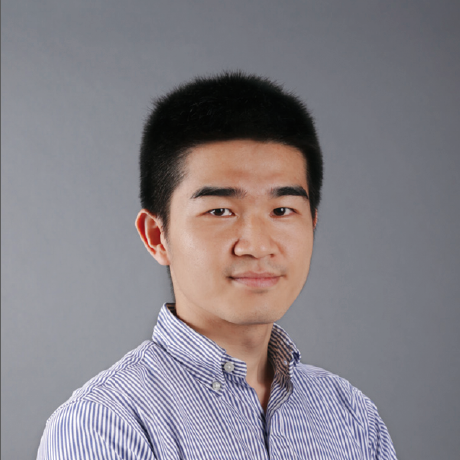}}]{Zhaopeng Tu}
is a Principal Researcher at the Tencent AI Lab, Shenzhen, China. He received his Ph.D. degree from the Institute of Computing Technology, Chinese Academy of Sciences in 2013. He was a Postdoctoral Researcher at the University of California at Davis from 2013 to 2014. He was a researcher at Huawei Noah’s Ark Lab, Hong Kong from 2014 to 2017. He is currently working on neural machine translation and Seq2Seq learning for other NLP tasks, such as dialogue and question answering.
\end{IEEEbiography}

\begin{IEEEbiography}[{\includegraphics[width=1in,height=1.25in,clip,keepaspectratio]{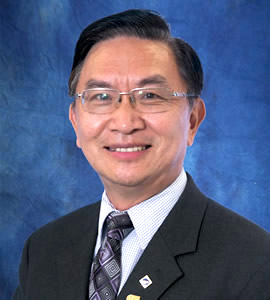}}]{Michael R. Lyu}
is currently a Professor at the Department of Computer Science and Engineering at The Chinese University of Hong Kong.
He received his B.S. in Electrical Engineering from National Taiwan University in 1981, his M.S. in Computer Science from University of California, Santa Barbara, in 1985, and his Ph.D. in Computer Science from University of California, Los Angeles in 1988.
His research interests include software engineering, dependable computing, distributed systems, cloud computing, mobile networking, big data, and machine learning.
He was elected to IEEE Fellow (2004), AAAS Fellow (2007), and ACM Fellow (2015) for his contributions to software reliability engineering and software fault tolerance.
\end{IEEEbiography}
\vfill

\end{document}